\documentclass[10pt,twocolumn,letterpaper]{article}

\usepackage{diagbox}
\usepackage{cvpr}
\usepackage{times}
\usepackage{epsfig}
\usepackage{graphicx}
\usepackage{amsmath}
\usepackage{amssymb}
\usepackage{color}
\usepackage{subfigure}

\usepackage[pagebackref=true,breaklinks=true,letterpaper=true,colorlinks,bookmarks=false]{hyperref}

\cvprfinalcopy % *** Uncomment this line for the final submission

\def\cvprPaperID{75} % *** Enter the CVPR Paper ID here

\ifcvprfinal\fi
\begin{document}

\hyphenpenalty=500

%%%%%%%%% TITLE
\title{Constrained Deep Metric Learning for Person Re-identification}

\author{Hailin Shi \and Xiangyu Zhu \and Shengcai Liao \and Zhen Lei \and Yang Yang \and Stan Z. Li \and
Institute of Automation, Chinese Academy of Sciences \\
{\tt \small \{hailin.shi,xiangyu.zhu,scliao,zlei,yangyang,szli\}@nlpr.ia.ac.cn}}

\maketitle
%\thispagestyle{empty}

%%%%%%%%% ABSTRACT
\begin{abstract}
   Person re-identification aims to re-identify the probe image from a given set of images under different camera views. It is challenging due to large variations of pose, illumination, occlusion and camera view.
   Since the convolutional neural networks (CNN) have excellent capability of feature extraction, certain deep learning methods have been recently applied in person re-identification.
   However, in person re-identification, the deep networks often suffer from the over-fitting problem.
   In this paper, we propose a novel CNN-based method to learn a discriminative metric with good robustness to the over-fitting problem in person re-identification.
   Firstly, a novel deep architecture is built where the Mahalanobis metric is learned with a weight constraint.
   This weight constraint is used to regularize the learning, so that the learned metric has a better generalization ability.
   Secondly, we find that the selection of intra-class sample pairs is crucial for learning but has received little attention.
   To cope with the large intra-class variations in pedestrian images, we propose a novel training strategy named moderate positive mining to prevent the training process from over-fitting to the extreme samples in intra-class pairs.
   Experiments show that our approach significantly outperforms state-of-the-art methods on several benchmarks of person re-identification.
\end{abstract}

%%%%%%%%% BODY TEXT
\section{Introduction}
\label{section_Introduction}

Given a set of pedestrian images, person re-identification aims to identify the probe image that generally captured by different cameras. Nowadays, person re-identification becomes increasingly important for surveillance and security system, \eg replacing manual video screening and other heavy loads.
Person re-identification is a challenging task due to large variations of body pose, lighting, view angles, scenarios across time and cameras.

The framework of existing methods usually consists of two parts: (1) extracting discriminative features from pedestrian images; (2) computing the distance of image pairs by feature comparison.
There are many works focus on these two aspects.
The traditional methods work at improving suitable hand-crafted features~\cite{yang2014salient, zhang2014novel, zhao2014learning}, or good metric for comparison~\cite{koestinger2012large, li2013locally, li2013learning, martinel2014saliency, zhao2013person}, or both of them~\cite{khamis2014joint, liao2015person, xiong2014person}. The first aspect considers to find features that are robust to challenging factors (lighting, pose \etc) while preserving the identity information. The second aspect comes to the metric learning problem which generally minimizes the intra-class distance while maximizing the inter-class distance.

\begin{figure}[!htb]
  \centering
  \includegraphics[width=0.35\textwidth]{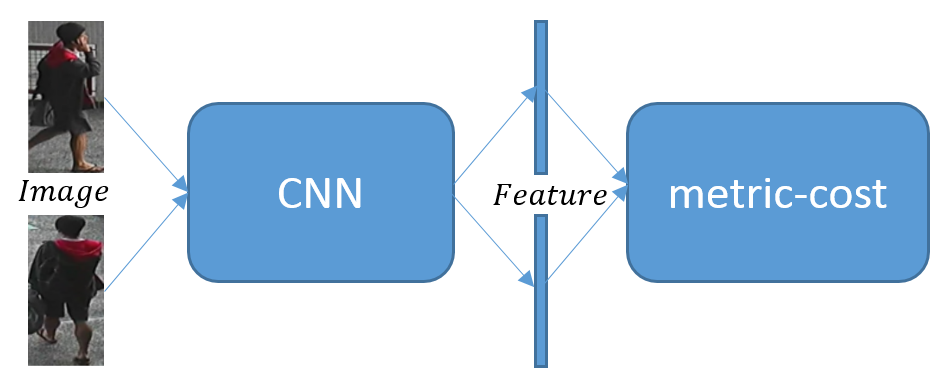}
  \caption{The general framework of deep learning methods for person re-identification.}
  \label{general-framework-DL}
\end{figure}

More recently, the deep learning methods gradually gain popularity in person re-identification.
The re-identification methods by deep learning~\cite{ahmed2015improved, ding2015deep, li2014deepreid, yi2014deep} incorporate the two above-mentioned aspects (feature extraction and metric learning) of person re-identification into an integrated framework.
The feature extraction and the metric learning are fulfilled respectively by two components in a deep neural network: (1) the CNN part which extracts features from pedestrian images, and (2) the following metric-cost part which compares the feature vectors with the chosen metric, computes the loss function, and back-propagates the gradient (Fig.~\ref{general-framework-DL}).
The FPNN~\cite{li2014deepreid} algorithm introduced a patch matching layer for the CNN part for the first time. Ahmed \etal~\cite{ahmed2015improved} proposed an improved deep learning architecture (IDLA) with cross-input neighborhood differences and patch summary features. These two methods are both dedicated to improve the CNN architecture.
Their purpose is to evaluate the pair similarity early in the CNN stage, so that it could make use of spatial correspondence of feature maps.
As for the metric-cost part, DML~\cite{yi2014deep} adopted the cosine similarity and Binomial deviance.
DeepFeature~\cite{ding2015deep} adopted the Euclidean distance and triplet loss.
Some others~\cite{ahmed2015improved, li2014deepreid} used the logistic loss to directly form a binary classification problem of whether the input image pair belongs to the same identity.

However, in person re-identification, the available training data is usually insufficient, which causes a weak generalization ability of existing deep learning methods on test data. To address this, in this paper, we propose a novel method of deep metric learning, and try two useful constraints to prevent the training from over-fitting. Specifically,
\begin{itemize}\itemsep=-1pt
    \item   The proposed network extracts features from a pair of images with a convolutional neutral network, and compares the two feature vectors with the Mahanalobis metric layers.
    The Mahanalobis metric layers are regularized by a weight constraint, so that the learned metric has a better generalization ability.
    The feature extractor and the metric layers are learned jointly.
    During the test, the network reads a pair of images and directly outputs the distance.
    \item   For training the deep neural network, the hard negative mining strategy has been commonly used~\cite{ahmed2015improved, parkhi2015deep, schroff2015facenet}.
    Considering the large intra-class variations in pedestrian data, we argue that, in person re-identification,
    the positive pairs should also be sampled carefully since forcing the model to deal with the extremely hard positives may cause over-fitting.
    This is an important issue but has been seldom noticed.
    In this paper, we propose a new training strategy, named moderate positive mining, to adaptively search the moderate positives for training and avoid the outliers.
    This novel strategy alleviates the over-fitting problem and significantly improves the identification accuracy.
\end{itemize}

\section{Related work}
\label{section_Related_work}

\paragraph{Constrainted metric learning.}
To our knowledge, there are seldom application of Mahalanobis metric in deep learning methods for person re-identification.
A commonly used metric by deep learning methods is the Euclidean distance.
However, the Euclidean distance is sensitive to the scale, and is blind to the correlation across dimensions.
In practice, we cannot guarantee that the CNN-learned features have similar scales and the de-correlation across dimensions.
Therefore, our method adopts the Mahalanobis distance which is a better choice of multivariate metric~\cite{manly2004multivariate}.
In the area of face recognition, DDML~\cite{hu2014discriminative} implemented the Mahalanobis distance in the network, but with hand-crafted features as input.
This is a significant difference with ours. We integrate the feature extraction and the Mahalanobis metric learning in a unified network, in which the two components are learned jointly.
Besides, our Mahalanobis metric is learned under a weight constraint (see Section~\ref{section_Weight_constraint}), which helps to gain a better generalization ability.
FaceNet~\cite{schroff2015facenet} and DeepFace~\cite{parkhi2015deep} implemented a similar metric in their networks, but without any weight constraint like ours.

\paragraph{Sample mining.}
The hard negative mining strategy~\cite{schroff2015facenet} is used more and more commonly for training deep networks.
In person re-identification, IDLA~\cite{ahmed2015improved} adopted hard negative mining in its training process.
By forcing the model to focus on the hard negatives near the decision boundary, hard negative mining improves the training efficiency and the model performance.
In this paper, we find that how to select moderate positive samples is also an essential issue for training person re-identification networks.
The moderate positives are as critical as hard negatives for training the network. However, there are barely any previous attempt in this aspect.
In our approach, we propose the novel strategy of moderate positive mining.
We sample the moderate positives for training, and avoid using outliers from extreme intra-class variations of pedestrian data.
We empirically find that this strategy effectively alleviates the over-fitting problem and improves the identification accuracy (see Section~\ref{section_Analysis_of_moderate_positive_mining}).

\paragraph{Branching Schemes in CNN.}
We build the CNN in the form of 3 ``branches", each of which is in charge of a fixed part of the input image (see Section~\ref{section_CNN_architecture_details} for details).
DML~\cite{yi2014deep} has a similar architecture compared with ours.
However, DML adopted weight sharing (\ie tied weights) between the branches, while ours does not. In Section~\ref{section_Filter_visualization}, we will show that untied branches, which learn more specific features from each part, are able to achieve better performance.

\section{Constrained Deep Metric Learning}
\label{section_Deep_Mahalanobis_Network}

\begin{figure*}[!htbp]
  \centering
  \includegraphics[width=0.9\textwidth]{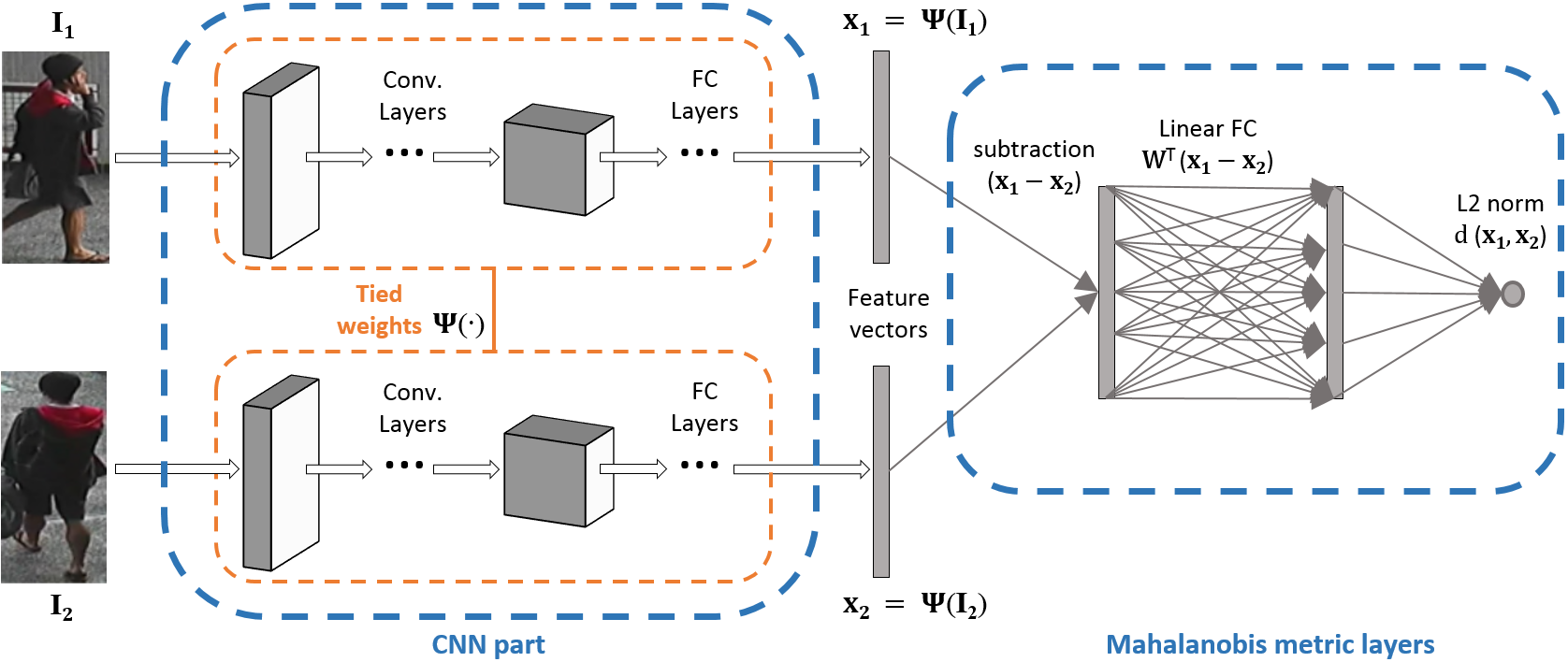}
  \caption{The overview of CDML Network. Best viewed in color.}
  \label{fig_overview}
\end{figure*}

The goal is to extract the features from two pedestrian images and compute their similarity with a discriminative metric. To achieve good performance, the image pairs of the same identity should have small distances, while those from different identities should have large distances.
In this work, we employ the convolutional neutral network, which has been proved to be of excellent capability to extract useful information from large-variation images~\cite{zeiler2014visualizing}. Fig.~\ref{fig_overview} is an overview of the network for Constrained Deep Metric Learning (CDML). The network can be divided into two parts, \ie the \emph{CNN part} and the \emph{Mahalanobis metric layers} from left to right in Fig.~\ref{fig_overview}. The first part extracts the features from the pedestrian images with two Siamese CNNs that share the weights (see Section~\ref{section_CNN_architecture_details} for architecture details). The second part is the Mahanalobis metric layers which aim to minimize the intra-class distance and maximize the inter-class distance. By incorporating the metric learning into the CNN framework, both feature extraction part and metric learning part can be trained jointly by gradient descent method, where the discriminability of both tasks can be improved.
Besides, the proposed weight constraint and the moderate positive mining strategy are employed to deal with the over-fitting problem.

\subsection{Mahalanobis metric layers}
\label{section_Mahalanobis_metric_layers}

Given two sets of pedestrian images $\mathcal{I}_\mathbf{1}$ and $\mathcal{I}_\mathbf{2}$ that come from two disjoint cameras,
$\mathcal{X}_\mathbf{1}$ and $\mathcal{X}_\mathbf{2}$ are the corresponding feature sets extracted by the CNN part.
Denote $\mathbf{x}_1\in\mathcal{X}_\mathbf{1}$ and $\mathbf{x}_2^p\in\mathcal{X}_\mathbf{2}$ as a positive pair (from the same identity), and $\mathbf{x}_1\in\mathcal{X}_\mathbf{1}$ and $\mathbf{x}_2^n\in\mathcal{X}_\mathbf{2}$ as a negative pair (from different identities).
The objective is to learn a Mahalanobis metric that minimizes the intra-class distance while maximizing the inter-class distance.
The Mahalanobis distance is formulated as
\begin{equation}\label{M-dist}
    d(\mathbf{x}_1, \mathbf{x}_2) = \sqrt{ (\mathbf{x}_1 - \mathbf{x}_2)^T \textbf{M} (\mathbf{x}_1 - \mathbf{x}_2) },
\end{equation}
where $\mathbf{x}_2 \in \{\mathbf{x}_2^p,\mathbf{x}_2^n\}$, $\textbf{M}$ is a symmetric positive semi-definite matrix.
In the traditional discriminative analysis problem where the features are known, the matrix $\textbf{M}$ can be solved under certain data distribution assumptions (\eg normal distribution).
In the framework of deep learning, however, the features $\mathbf{x}_1$ and $\mathbf{x}_2$ are unknown before the CNN is learned.
Therefore, it is natural to learn the matrix $\textbf{M}$ and the CNN \emph{jointly} by back-propagation.

Denote the $\mathbf{\Psi} (\cdot)$ as the front-end CNN, $\mathbf{I}_1 \in \mathcal{I}_\mathbf{1}$ and $\mathbf{I}_2 \in \mathcal{I}_\mathbf{2}$ as the corresponding images that $\mathbf{x}_1 = \mathbf{\Psi}(\mathbf{I}_1)$ and $\mathbf{x}_2 = \mathbf{\Psi}(\mathbf{I}_2)$.
Since the matrix $\textbf{M}$ is symmetric and positive semi-definite, we make use of its decomposition $\textbf{M} = \textbf{W}\textbf{W}^T$.
This is because directly learning $\textbf{M}$ under the constraint of positive semi-definite is difficult,
whereas learning $\textbf{W}$ is much easier, and $\textbf{W}\textbf{W}^T$ is always positive semi-definite.
We develop the distance as follows
\begin{align}\label{M-dist-dev-2}
    & d(\mathbf{x}_1, \mathbf{x}_2)  = \sqrt{ (\mathbf{\Psi}(\mathbf{I}_1) - \mathbf{\Psi}(\mathbf{I}_2))^T \textbf{W}\textbf{W}^T (\mathbf{\Psi}(\mathbf{I}_1) - \mathbf{\Psi}(\mathbf{I}_2)) }    \nonumber   \\
    & = \sqrt{ (\textbf{W}^T(\mathbf{\Psi}(\mathbf{I}_1) - \mathbf{\Psi}(\mathbf{I}_2)))^T(\textbf{W}^T(\mathbf{\Psi}(\mathbf{I}_1) - \mathbf{\Psi}(\mathbf{I}_2))) }            \nonumber   \\
    & = \|\textbf{W}^T(\mathbf{\Psi}(\mathbf{I}_1) - \mathbf{\Psi}(\mathbf{I}_2))\|_2.
\end{align}
The inner product $\textbf{W}^T(\mathbf{\Psi}(\mathbf{I}_1) - \mathbf{\Psi}(\mathbf{I}_2))$ can be implemented by a linear fully-connected (FC) layer in which the weight matrix is defined by $\textbf{W}^T$.
The output of the FC layer is calculated by
\begin{equation}\label{FC_compute}
    \mathbf{y} = f(\textbf{W}^T\mathbf{x} + \mathbf{b}),
\end{equation}
where $\mathbf{b}$ is the bias term.
The identity function is used as the activation $f(\cdot)$ for the linear FC layer.

Therefore, we implement the Mahalanobis metric in a neural network form (the right part in Fig.~\ref{fig_overview}) after the CNN.
First, the feature vectors $\mathbf{\Psi}(\mathbf{I}_1)$ and $\mathbf{\Psi}(\mathbf{I}_2)$ (\ie $\mathbf{x}_1$ and $\mathbf{x}_2$) extracted by CNN are fed into the subtraction unit.
Then, the difference is transformed by the linear FC layer with the weight matrix $\textbf{W}^T$.
For the symmetry of the distance, we fix the bias term $\mathbf{b}$ of the FC layer to zero throughout the training and test.
%Thus, the first order term and the constant term are annihilated.
Finally, the L2 norm is computed as the output distance $d(\mathbf{\Psi}(\mathbf{I}_1), \mathbf{\Psi}(\mathbf{I}_2))$.
This structure remains equivalent when switching the position of the subtraction unit and the FC layer.
The training loss is defined as
\begin{equation}\label{Loss}
    \mathbf{\emph{L}} = d(\mathbf{\Psi}(\mathbf{I}_1), \mathbf{\Psi}(\mathbf{I}_2^p)) - d(\mathbf{\Psi}(\mathbf{I}_1), \mathbf{\Psi}(\mathbf{I}_2^n)),
\end{equation}
where $\mathbf{I}_2^p$ and $\mathbf{I}_2^n$ are the input images corresponding to the features $\mathbf{x}_2^p$ and $\mathbf{x}_2^n$.
In each time of the forward propagation, either the first term or the second term of Eq.~\ref{Loss} is computed.
Then the training loss is obtained by combining the two terms, and we compute the gradient and back-propagate it.
Similar to the triplet loss~\cite{schroff2015facenet}, this training objective aims to minimize the positive distance and maximize the negative distance.

\subsection{Weight constraint}
\label{section_Weight_constraint}

As mentioned above, the Mahalanobis metric layers aim to learn a discriminative metric matrix $\textbf{M}$ for minimizing the intra-class distance and maximizing the inter-class distance.
Compared with the Mahalanobis distance, the Euclidean distance has less discriminability but better generalization ability, because it does not take account of the scales and the correlation across dimensions~\cite{manly2004multivariate}.
Here, we impose a constraint that keep the matrix $\textbf{M}$ having large values at the diagonal and small entries elsewhere, so we can achieve a balance between the unconstrained Mahalanobis distance and the Euclidean distance.
The constraint is formulated as the Frobenius norm of the difference between $\textbf{W}\textbf{W}^T$ and identity matrix $\mathbf{I}$,
\begin{align}\label{Loss-st}
    \mathbf{\emph{L}} = d(\mathbf{\Psi}(\mathbf{I}_1), \mathbf{\Psi}(\mathbf{I}_2^p)) - d(\mathbf{\Psi}(\mathbf{I}_1), \mathbf{\Psi}(\mathbf{I}_2^n))       \nonumber   \\
    s.t. \quad \|\textbf{W}\textbf{W}^T - \mathbf{I}\|^2_F \leq \emph{C},
\end{align}
where $\emph{C}$ is a constant. We further combine the constraint into the loss function as a regularization term:
\begin{equation}\label{Loss-lambda}
    \hat{ \mathbf{\emph{L}} } = \mathbf{\emph{L}} + \frac{\lambda}{2} \|\textbf{W}\textbf{W}^T - \mathbf{I}\|^2_F,
\end{equation}
where $\lambda$ is the relative weight of regularization, $\hat{ \mathbf{\emph{L}} }$ is the new loss function.
For updating the weight matrix $\textbf{W}$, the gradient w.r.t $\textbf{W}$ is computed by
\begin{equation}\label{Loss-gradient}
    \frac{\partial \hat{ \mathbf{\emph{L}} }}{\partial \textbf{W}} =
    \frac{\partial \mathbf{\emph{L}}}{\partial \textbf{W}} + \lambda (\textbf{W}\textbf{W}^T - \mathbf{I})\textbf{W}.
\end{equation}

When $\lambda$ is large, the matrix $\textbf{M}$ becomes close to the identity matrix.
In the extreme case, $\textbf{M}$ equals to the identity matrix, and the distance
degenerates to the Euclidean distance.
In this situation, the metric has low variance but high bias, because it does not take account of the scales and the correlation across dimensions.
In the other case where $\lambda$ is too small, the metric fits the training data well, but is endangered by over-fitting.
So, in the training, we make use of the weight constraint to alleviate the over-fitting by balancing the variance and bias.

\subsection{CNN with untied branches}
\label{section_CNN_architecture_details}

\begin{figure}[!htb]
  \centering
  \includegraphics[width=0.5\textwidth]{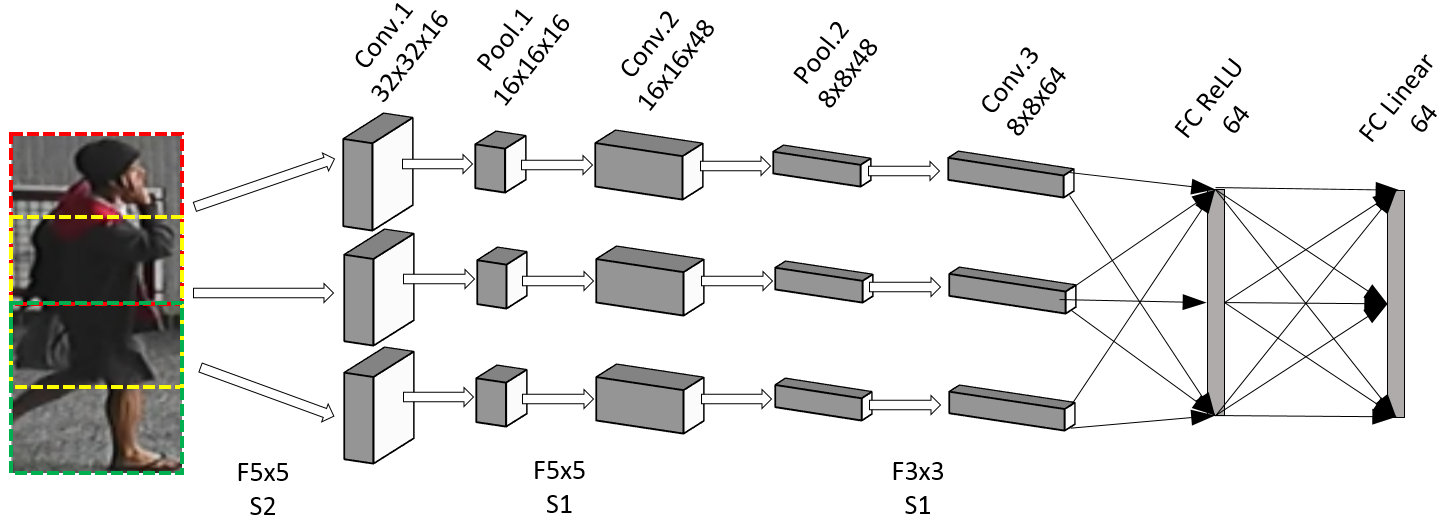}
  \caption{The CNN architecture we use for feature extraction. The 3 branches do not share weights with each other.
  Top: layer type and output size. Bottom: convolution parameters with "F" and "S" denoting the filter size and stride, respectively.}
  \label{cnn-module}
\end{figure}

In the beginning of this section, Fig.~\ref{fig_overview} roughly presents the Siamese CNNs with tied weights.
In fact, each CNN is built by 3 branches with the details shown in Fig.~\ref{cnn-module}.
The input image is firstly normalized to a $128\times64$ RGB image. Then, it is split into three $64\times64$ overlapping color patches, each of which is charged by a branch. Each branch is constituted of 3 convolutional layers and 2 pooling layers. No parameter sharing is performed between branches within a CNN. Then, the 3 branches are concluded by a FC layer with the ReLU activation. Finally, the output feature vector is computed by another FC layer with linear activation.
The replicate of this CNN extracts the feature vector from the other input image.
For the computational stability, the features are normalized before sending to the metric layers.
The proposed metric layers are performed subsequently to calculate the cost and gradient.

The reason that we build the CNN architecture in branches is to learn specific features from each part.
DML~\cite{yi2014deep} adopted a similar architecture but with tied weights between branches.
In Section~\ref{section_Filter_visualization}, the experiments show the advantage of our architecture.

\section{Moderate positive mining}
\label{section_Moderate_positive_mining}

\begin{figure}[!htb]
  \centering
  \includegraphics[width=0.45\textwidth]{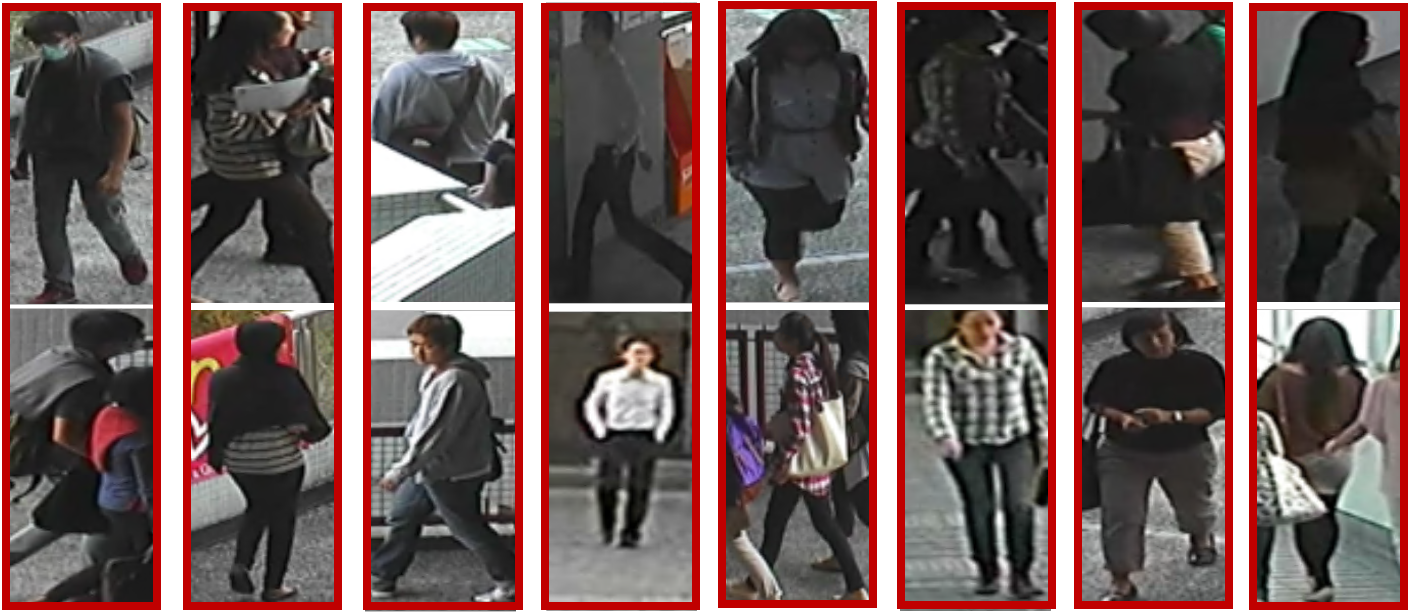}
  \caption{Some hard positive cases in CUHK03 labeled.}
  \label{hard-positives}
\end{figure}

There are many factors that lead to the large intra-class variations in pedestrian data, such as illumination, background, misalignment, occlusion, co-occurrence of people, appearance changing, \etc.
Many of them are specific with pedestrian data.
Fig.~\ref{hard-positives} shows some hard positive cases in the data set of CUHK03~\cite{li2014deepreid}.
Some of them are even difficult for human to recognize.
We argue that using these extremely hard positive pairs to train the network may harm the practical performance because if the network is forced to handle these hard positives, it has a very large possibility of over-fitting.

As described in Section~\ref{section_Weight_constraint}, we propose the weight constraint to alleviate the over-fitting problem.
However, to deal with the over-fitting to the bad samples in positive pairs, only regularizing the metric layer weights is insufficient.
We need a better strategy for the selection of positive pairs.
Therefore, we introduce the moderate positive mining method as follows: we select the moderate positive pairs in the range of the same subject at one time.
For example, suppose a subject having 6 images, of which 3 from a camera and 3 from another.
We can totally match 9 positive pairs from this subject.
If we use the easiest or hardest positive pairs of the nine, the training will be very slow, and the network will be biased.
Thus, we pick the moderate positive pairs that are between the two extreme cases.
The mining criterion is described as follows:
\begin{equation}\label{Loss-epm}
     \alpha \leq \frac{d(\mathbf{\Psi}(\mathbf{I}_1), \mathbf{\Psi}(\hat{\mathbf{I}}_2^p)) - \min_{\mathbf{I}_2^p}{d(\mathbf{\Psi}(\mathbf{I}_1), \mathbf{\Psi}(\mathbf{I}_2^p))}}
     {\max_{\mathbf{I}_2^p}{d(\mathbf{\Psi}(\mathbf{I}_1), \mathbf{\Psi}(\mathbf{I}_2^p))} - d(\mathbf{\Psi}(\mathbf{I}_1), \mathbf{\Psi}(\hat{\mathbf{I}}_2^p))} \leq \beta,
\end{equation}
where $\alpha$ and $\beta$ are non-negative, $\hat{\mathbf{I}}_2^p$ is the selected image satisfying the mining criterion. %Normally, we assign $\alpha$ a small value for the easy positive mining.
The difficulty level increases as $\alpha$ and $\beta$ increase, and decreases otherwise.

\section{Experiments}
\label{section_Experiments}

Our network is implemented using the CUDA-Convnet~\cite{krizhevsky2012imagenet} framework.
We report the standard evaluation on three common benchmarks of person re-identification, \ie CUHK03~\cite{li2014deepreid}, CUHK01~\cite{li2012human} and VIPeR~\cite{gray2007evaluating}.
The proposed method is compared with state of the art on each data set.
All the evaluations are reported in the single-shot setting.
We begin with the experiments on CUHK03 in both labeled and detected version.
CUHK03 is a large data set which is suitable for performing deep networks.
We then analyze the effects of the untied branches, the moderate positive mining strategy and the weight constraint.
Finally, we evaluate our approach on the small data sets CUHK01 and VIPeR.

\subsection{Implementation details}
\label{section_Initialization}

Since the proposed method has a deep architecture, the initialization of parameters becomes crucial.
We initialize the CNN part and the Mahalanobis metric layers separately.
For the experiments on CUHK03, we initialize the CNN by pre-training it with softmax classification on the training set of CUHK03.
The outputs of softmax correspond to the person identities. Afterwards, we discard the softmax layer, and keep the pre-trained CNN as its initialization.
As for the metric layers, we initialize the weight matrix $\textbf{W}$ by a $64 \times 64$ identity matrix.

For the experiments on CUHK01 and VIPeR, we encounter the problem of small training set.
To exploit the advantage of deep learning, we use a large data set to initialize the CNN which is further fine-tuned on the small data sets (the same as the training strategy of IDLA\cite{ahmed2015improved}).
To our knowledge, Market1501~\cite{zheng2015scalable} is currently the largest public data set of person re-identification. It has totally 1,501 subjects, each of which has around 22 images.
We utilize the entire data set of Market1501 and CHUK03 to implement the softmax pre-training of CNN.
We fine-tune the whole deep network on the training sets of CUHK01 and VIPeR, and evaluate it on their test sets, respectively.

We set the parameter $\lambda = 10^{-2}$ in all the following experiments except the analysis of $\lambda$ itself in Section~\ref{section_Analysis_of weight_constraint}.
The parameters $\alpha$ and $\beta$ are determined adaptively according to the distribution of positive samples.

\subsection{Experiments on CUHK03}
\label{section_Experiments_on_CUHK03}

CUHK03 contains 1,369 subjects, each of which has around 10 images.
The default protocol randomly selects 1,160 subjects for training, 100 for validation, and 100 for test.
We adopt a random translation for the training data augmentation.
Our pre-trained network is fine-tuned on the training set in a pair-wise way.
Note that, for the experiments on CHUK03, both the pre-training and the fine-tuning are done on the training set of CUHK03.
The proposed weight constraint and moderate positive mining are employed. Besides, the hard negative mining is also used in the training.

\begin{figure*}[!htbp]
  \centering
  \subfigure[labeled]{
  \includegraphics[width=0.49\textwidth]{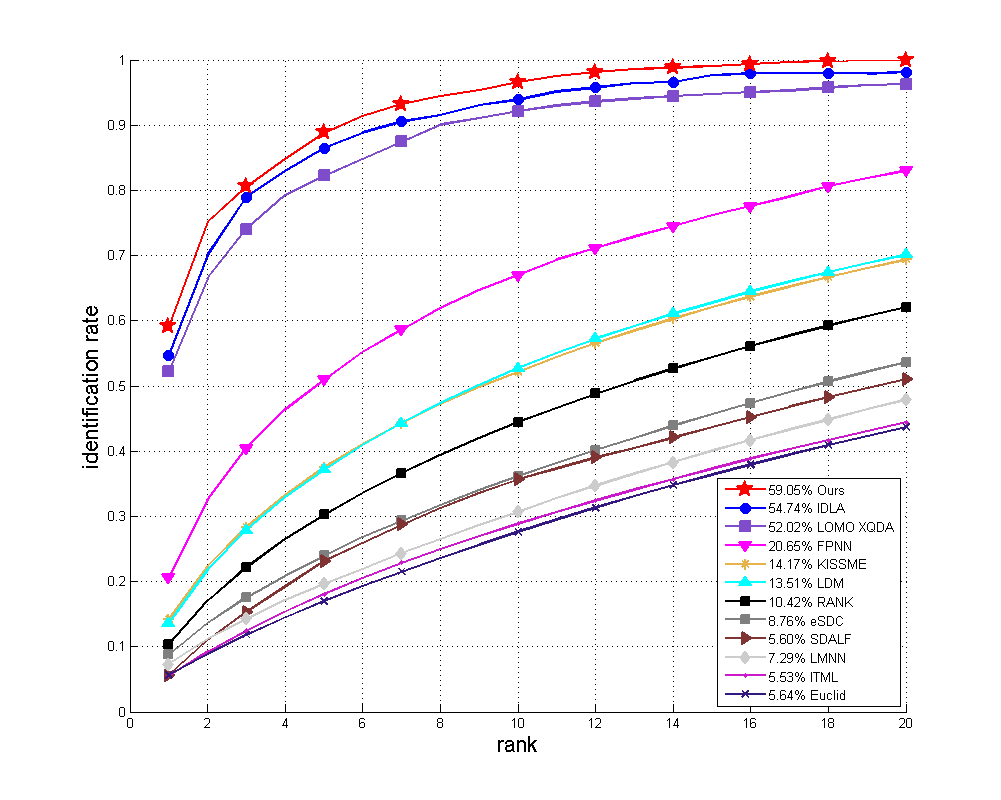}}
  \subfigure[detected]{
  \includegraphics[width=0.49\textwidth]{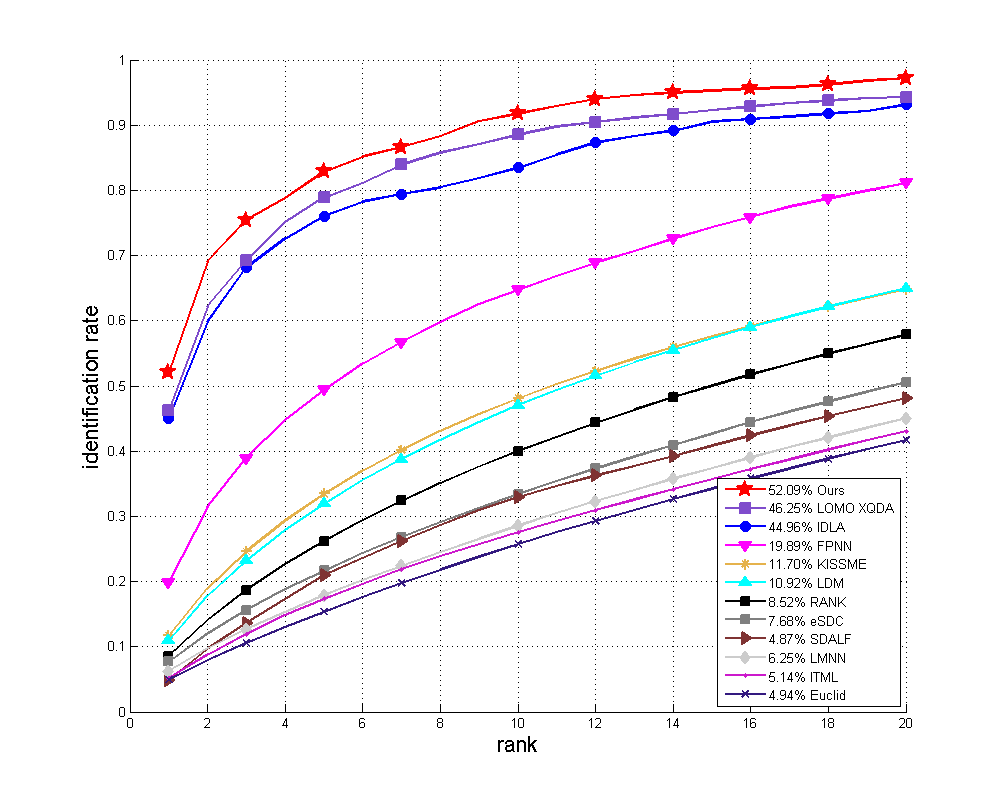}}

  \caption{CMC curves and rank-1 identification rates on the CHUK03 data set. Our method outperforms the previous methods on both labeled (a) and detected (b) versions.}
  \label{fig_cmc_cuhk03}
\end{figure*}

CUHK03 has 2 versions, one is manually labeled images, and the other is detected images. We evaluate our method on both versions.
We compare our performance with the traditional methods and deep learning methods. The traditional methods include LOMO-XQDA~\cite{liao2015person}, KISSME~\cite{koestinger2012large}, LDM~\cite{guillaumin2009you}, RANK~\cite{mcfee2010metric}, eSDC~\cite{zhao2013unsupervised}, SDALF~\cite{farenzena2010person}, LMNN~\cite{weinberger2005distance}, ITML~\cite{davis2007information}, Euclid~\cite{zhao2013unsupervised}. The deep learning methods include FPNN~\cite{li2014deepreid} and IDLA~\cite{ahmed2015improved}. IDLA and LOMO-XQDA gained the previously best performance on CUHK03.
The cumulative matching characteristic (CMC) curves and the rank-1 identification rates are shown in Fig.~\ref{fig_cmc_cuhk03}.
Our method achieves better performance than the previous state-of-the-art methods on not only the labeled version but also the detected version.
This indicates that our method achieves good robustness to the misalignment of detection.

\subsection{Analysis of untied branches}
\label{section_Filter_visualization}

\begin{figure}[!htb]
  \centering
  \includegraphics[width=0.46\textwidth]{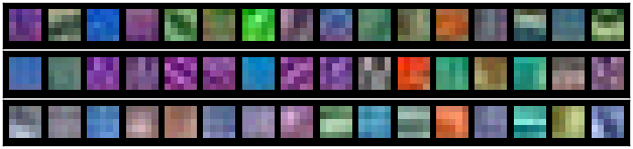}
  \caption{The learned filters of the first convolutional layer. The top, middle and bottom line correspond to the top, middle and bottom branches in the proposed CNN, respectively. Best viewed in color.}
  \label{fig_filters_cuhk03}
\end{figure}

\begin{figure}[!htb]
  \centering
  \includegraphics[width=0.36\textwidth]{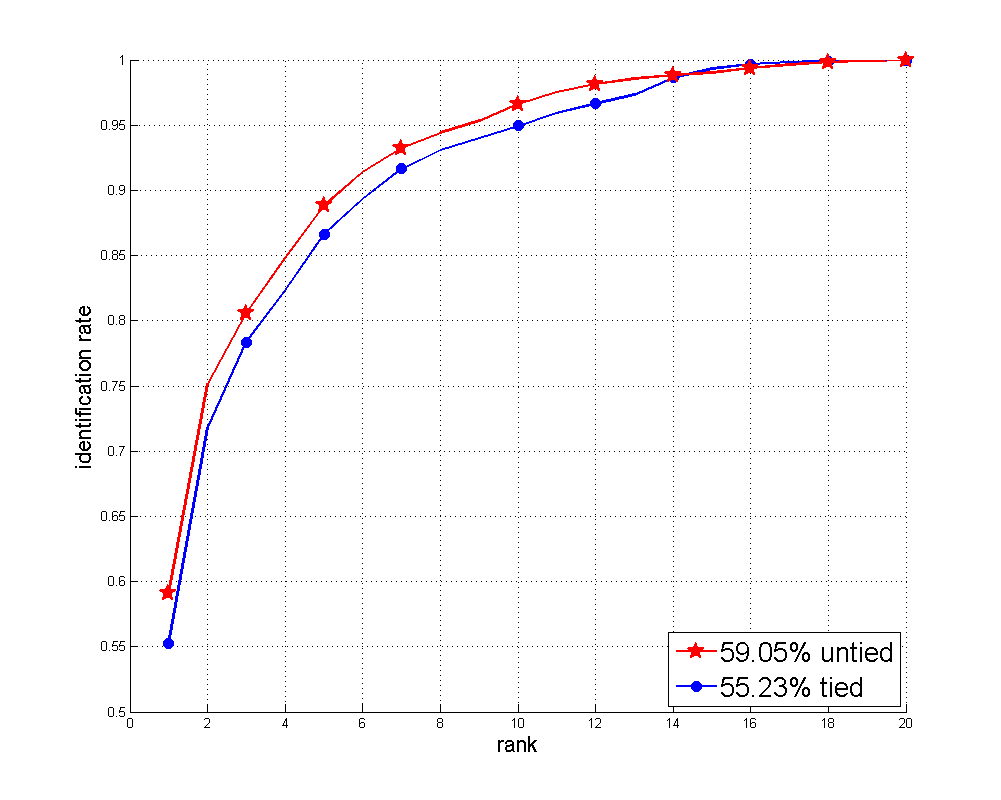}
  \caption{The performances with and without tied weights between branches on CUHK03 labeled.}
  \label{fig_cmc_cuhk03_tied}
\end{figure}

We show the learned filters of untied branches in Fig.~\ref{fig_filters_cuhk03}.
We find that the network has learned remarkable color representations, which is coherent with results of IDLA~\cite{ahmed2015improved}.
Since we do not apply tied weights between branches, each branch learns different filters from their own part.
As shown in Fig.~\ref{fig_filters_cuhk03} where each row demonstrates a filter set from one branch,
we can find that each branch has its own emphasis in color.
For example, the middle branch inclines to violet and blue, whereas the bottom branch has learned filters of obviously lighter colors than the other 2 branches.
The reason is that the different parts of pedestrian image have different color distributions.
Therefore, the branches learn the part-specific filters.
We compare the performances with and without tied weights between branches in Fig.~\ref{fig_cmc_cuhk03_tied}.
The untied-branch network gains a better performance than that of tied branches.

\subsection{Analysis of moderate positive mining}
\label{section_Analysis_of_moderate_positive_mining}

\begin{figure}[!htb]
  \centering
  \includegraphics[width=0.36\textwidth]{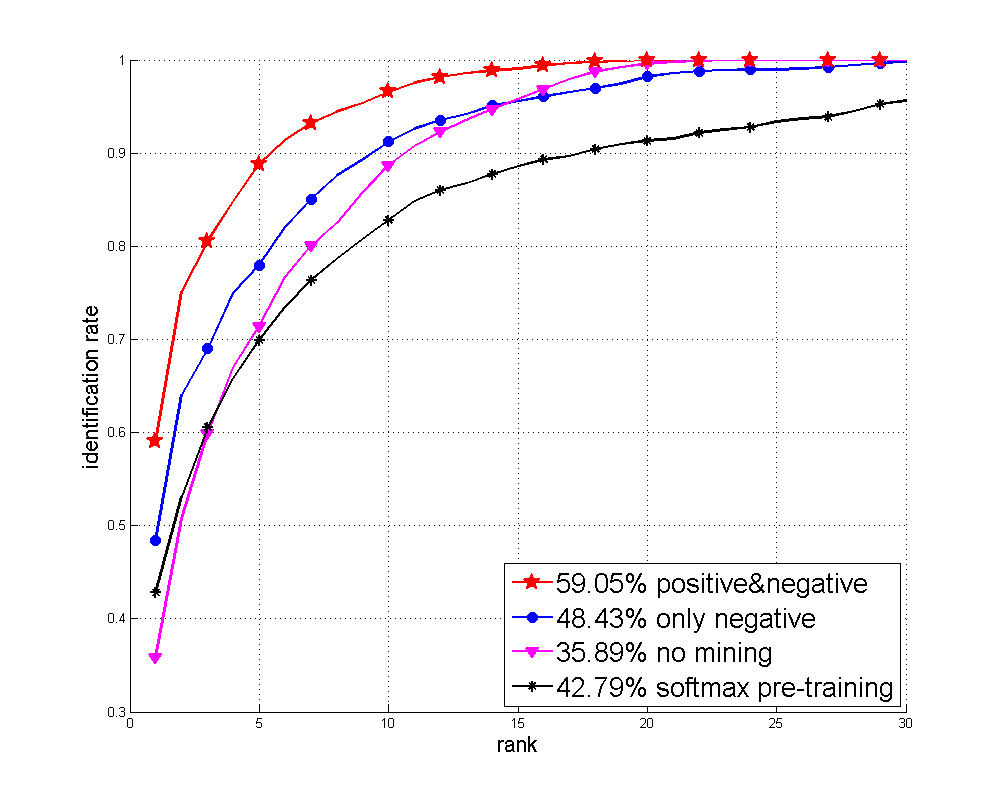}
  \caption{Performance comparison of moderate positive mining on CUHK03 labeled.
  Red: both moderate positive mining and hard negative mining are employed.
  Blue: only hard negative mining is employed.
  Magenta: no mining technique is employed during training.
  Black: the softmax pre-trained network.}
  \label{fig_epm_compare}
\end{figure}

In the above experiments, we employ both the proposed moderate positive mining and hard negative mining in the training.
To further demonstrate the advantage of moderate positive mining, we compare the performances with and without the moderate positive mining.
We also compare them with the pre-trained network.
Their CMC curves and rank-1 identification rates are shown in Fig.~\ref{fig_epm_compare}.
From the CMC curves, we can find that the collaboration of moderate positive mining and hard negative mining achieves the best result (red line).
The absence of moderate positive mining leads to a significant derogation of performance (blue).
If both of the two mining methods are not used (magenta), the network gives very low identification rate at low ranks, even worse than the pre-trained network (black).
This indicates that moderate positive mining and hard negative mining are both crucial for training.

The CMC curves of the 3 trained networks tend to converge after the rank exceeds 20, whereas the pre-trained network remains at a relatively low identification rate.
This indicates that the training with the Mahalanobis metric layers is the basic contributor of the improvement.

\subsection{Analysis of weight constraint}
\label{section_Analysis_of weight_constraint}

\begin{figure}[!htb]
  \centering
  \includegraphics[width=0.36\textwidth]{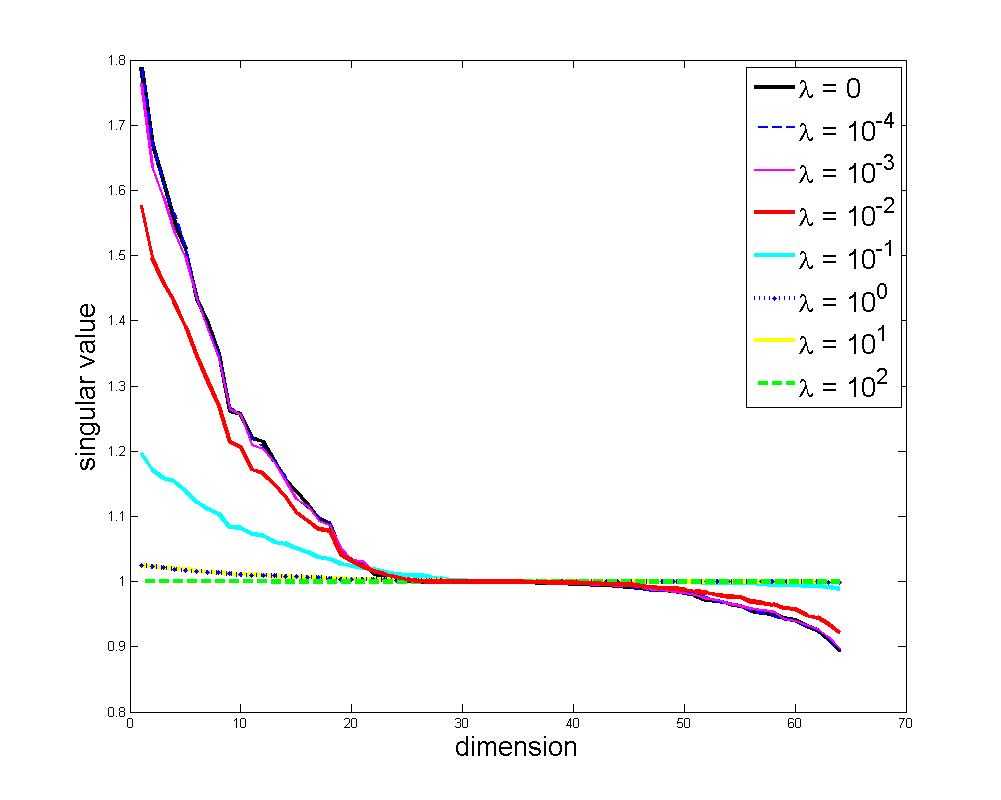}
  \caption{The spectrums of the matrix $\textbf{M}$. The spectrums with $\lambda=10^1, 10^0$ are very close; those with $\lambda=10^{-3}, 10^{-4}, 0$ are also very close. Best viewed in color.}
  \label{fig_lambda_spectrum}
\end{figure}

\begin{figure}[!htb]
  \centering
  \includegraphics[width=0.3\textwidth]{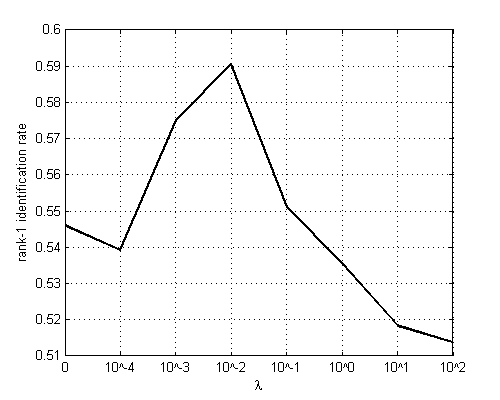}
  \caption{The rank-1 identification rates on CUHK03 labeled with different $\lambda$ of the weight constraint.}
  \label{fig_lambda_rank1}
\end{figure}

For preventing the Mahalanobis metric layers from over-fitting, we use the weight constraint as a regularization term.
Here, we analyze the metric matrices learned with different relative weights ($\lambda$) of the regularization.
In Fig.~\ref{fig_lambda_spectrum}, we show the spectrums of the matrix $\textbf{M}$.
We also show the corresponding rank-1 identification rates in Fig.~\ref{fig_lambda_rank1}.

When $\lambda = 10^2$, the singular values are almost constant at 1, which means the metric layers almost give the Euclidean distance. This leads to the low variance and high bias (see Section~\ref{section_Weight_constraint}).
As $\lambda$ increases, the matrix has varying singular values across dimensions.
This implies that the learned metric suits the training data well, but is more likely to have over-fitting.
Therefore, a moderate value of $\lambda$ gives a trade-off between the variance and bias, which is an appropriate choice for good performance (Fig.~\ref{fig_lambda_rank1}).

\subsection{Experiments on CUHK01}
\label{section_Experiments_on_CUHK01}

\begin{figure}[!htb]
  \centering
  \includegraphics[width=0.49\textwidth]{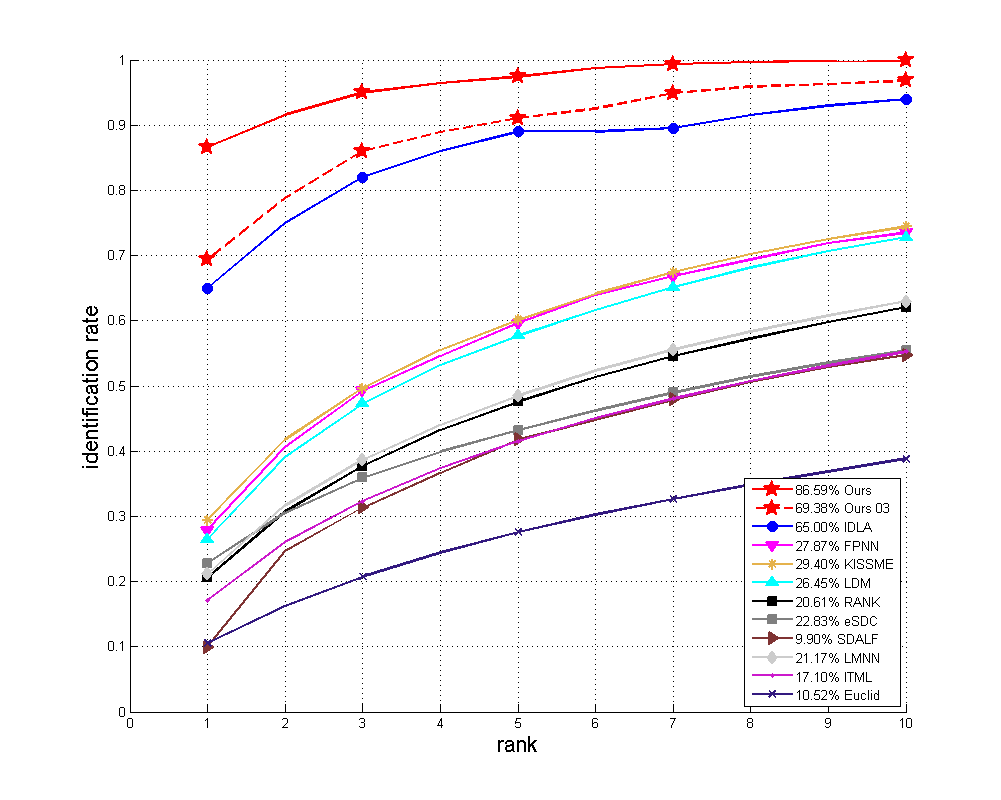}
  \caption{CMC curves and rank-1 identification rates on CHUK01.}
  \label{fig_cmc_cuhk01}
\end{figure}

The CUHK01 data set contains 971 subjects, each of which has 4 images under 2 camera views.
According to the protocol in ~\cite{li2012human}, the data set is divided into a training set of 871 subjects and a test set of 100.
As described in Section~\ref{section_Initialization}, we pre-train the CNN on Market1501 and CUHK03, and fine-tune the whole network on the training set of CUHK01.
The moderate positive mining and hard negative mining are employed.
We compare our approach with the previously mentioned methods. The CMC curves and rank-1 identification rates are shown in Fig.~\ref{fig_cmc_cuhk01}.
Our approach outperforms the state-of-the-art method (IDLA~\cite{ahmed2015improved}) by a large gap (with the rank-1 identification rate rising from 65\% to 87\%).
Besides, for a fair comparison with IDLA, we use only CUHK03 to pre-train the CNN, and use CUHK01 for fine-tuning (as the same setting in IDLA).
Under this setting, our approach also achieves better performance (marked as ``\emph{Ours 03}'' in Fig.~\ref{fig_cmc_cuhk01}) than IDLA.

\begin{figure*}[!htb]
  \centering
  \includegraphics[width=0.9\textwidth]{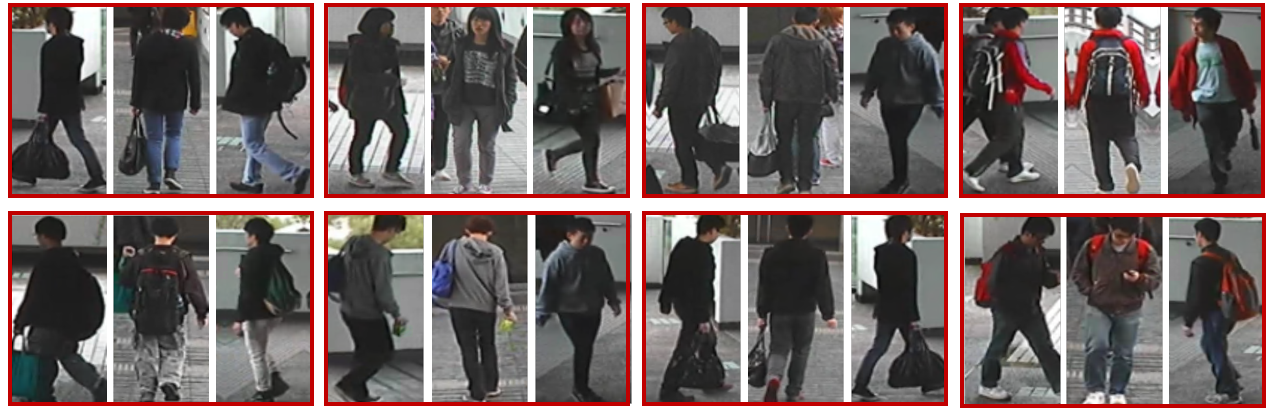}
  \caption{Some failed cases on CHUK01 by the proposed method. Left: true gallery. Middle: probe. Right: false positive.}
  \label{fig_cuhk01_failed}
\end{figure*}

To inspect the limitation on CUHK01, we show some failed cases in Fig.~\ref{fig_cuhk01_failed}. In each block, we give the true gallery, probe and false positive image from left to right.
We find that most failed cases come from the dark color images or the negative pairs of significant color correspondence.
This phenomenon is in line with the fact~\cite{ahmed2015improved} that the learned filters in network mainly focus on image colors (as shown in Fig.~\ref{fig_filters_cuhk03}).
The re-identification problem becomes extremely difficult when the true positive pairs have inconsistent colors in view while the negative pairs have similar colors (due to the lighting, camera setting \etc).

\subsection{Experiments on VIPeR}
\label{section_Experiments_on_VIPeR}

\begin{figure}[!htb]
  \centering
  \includegraphics[width=0.49\textwidth]{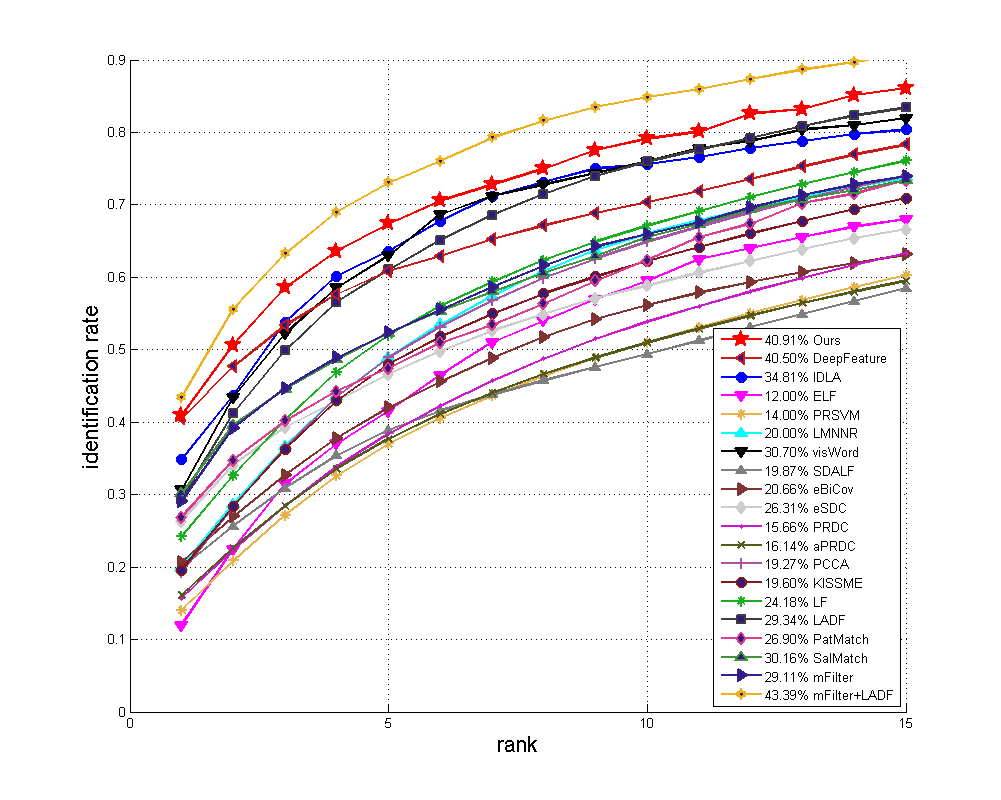}
  \caption{CMC curves and rank-1 identification rates on VIPeR.}
  \label{fig_cmc_viper}
\end{figure}

The VIPeR~\cite{gray2007evaluating} data set includes 632 subjects, each of which has 2 images from two different cameras.
Although VIPeR is a small data set which is not suitable for training CNN, we are still interested in the performance on this challenging task.
The data set is randomly split into two subsets, each has non-overlapping subjects of the same size. The two subsets are for either training or test.
We fine-tune the network on the 316-person training set and test it on the test set.
We also adopt a random translation for training data augmentation.
The moderate positive mining and hard negative mining are both used.
The results are shown in Fig.~\ref{fig_cmc_viper}.
We compare our model with IDLA~\cite{ahmed2015improved}, DeepFeature~\cite{ding2015deep}, visual word (visWord)~\cite{zhang2014novel}, saliency matching (SalMatch), patch matching (PatMatch)~\cite{zhao2013person}, ELF~\cite{gheissari2006person}, PRSVM~\cite{bazzani2012multiple}, LMNNR~\cite{bak2011multiple}, eBiCov~\cite{ma2012bicov}, local Fisher discriminant analysis (LF)~\cite{pedagadi2013local}, PRDC~\cite{zheng2011person}, aPRDC~\cite{liu2012person}, PCCA~\cite{mignon2012pcca}, mid-level filters (mFilter)~\cite{zhao2014learning} and the fusion of mFilter and LADF~\cite{li2013learning}.
Our approach achieves the identification rate of 40.91\% at rank 1, which is the best result on VIPeR compared with the existing deep learning methods.
Note that the highest rank-1 identification rate (43.39\%) is obtained by a combination of two methods (mFilter+LADF)~\cite{li2013learning}.
The identification rate by DeepFeature~\cite{ding2015deep} is close to ours at rank 1, but much lower at higher ranks.

\section{Conclusion}
\label{section_Conclusion}

In this paper, we propose the Constrained Deep Metric Learning method to learn a discriminative metric with good robustness to the over-fitting problem in person re-identification.
The Mahalanobis metric layers are regularized by the weight constraint, so that the learned metric has a good generalization ability.
The network learns the CNN feature extractor and the Mahalanobis metric layers jointly.
Moreover, we find that the selection of positive samples for training deep networks is as important as the negatives in the re-identification task.
Accordingly, we propose a new training strategy with moderate positive mining, which selects moderate positive pairs for training and so prevents the network from over-fitting to the bad positives.
Due to these improvements, our method achieves state-of-the-art performances on the data sets of CUHK03, CUHK01, and VIPeR compared with other deep learning methods.

\section{Acknowledgments}
This work was supported by the Chinese National Natural Science Foundation Projects \#61105023, \#61103156, \#61105037, \#61203267, \#61375037, \#61473291, National Science and Technology Support Program Project \#2013BAK02B01, Chinese Academy of Sciences Project No.KGZD-EW-102-2, and AuthenMetric R\&D Funds.
The Tesla K40 used for this research was donated by the NVIDIA Corporation.

{\small
\bibliographystyle{ieee}
\bibliography{epm}
}

\end{document}